\documentclass{article}
\usepackage{ijcai19}

\usepackage{times}
\usepackage{soul}
\usepackage{url}
\usepackage[utf8]{inputenc}
\usepackage[small]{caption}
\usepackage{graphicx}
\usepackage{amsmath}
\usepackage{booktabs}
\usepackage{subfig}

\newcommand{\T}{\mathcal{P}}
\newcommand{\R}{\mathcal{R}}
\renewcommand{\S}{\mathcal{S}}
\newcommand{\A}{\mathcal{A}}

\usepackage{fancyhdr}

\title{Leveraging Human Guidance for Deep Reinforcement Learning Tasks}

\author{
Ruohan Zhang\footnote{Contact Author} \and Faraz Torabi \and Lin Guan \and Dana H. Ballard \And Peter Stone
\affiliations
Department of Computer Science, University of Texas at Austin, USA\\
\emails
\{zharu, guanlin, danab\}@utexas.edu; \{faraztrb, pstone\}@cs.utexas.edu}

\begin{document}
\thispagestyle{fancy}
\maketitle

\begin{abstract}
Reinforcement learning agents can learn to solve sequential decision tasks by interacting with the environment. Human knowledge of how to solve these tasks can be incorporated using imitation learning, where the agent learns to imitate human demonstrated decisions. However, human guidance is not limited to the demonstrations. Other types of guidance could be more suitable for certain tasks and require less human effort. This survey provides a high-level overview of five recent learning frameworks that primarily rely on human guidance other than conventional, step-by-step action demonstrations. We review the motivation, assumption, and implementation of each framework. We then discuss possible future research directions. 
\end{abstract}

\section{Introduction}
In end-to-end learning of sequential decision tasks, algorithms such as imitation learning (IL) \cite{osa2018algorithmic}, reinforcement learning (RL) \cite{sutton2018reinforcement}, or a combination of both \cite{silver2016mastering} have achieved remarkable successes in many practical problems. These algorithms enable learning agents to learn an optimal \emph{policy}
through imitating a human demonstrator's behaviors (IL), or through trial and error by interacting with the environment (RL). Recent advancement in deep learning has enabled these learning algorithms to solve more challenging tasks~\cite{mnih2015human,silver2016mastering}. For these tasks, one major issue with deep learning algorithms is their sample efficiency. For instance, an RL agent may require millions of training samples to learn a good policy to play a video game~\cite{mnih2015human}. In practice different types of human guidance are often introduced as sources of domain knowledge to speed up learning. 

The most common form of human guidance is the human \emph{policy} itself. A human trainer communicates the policy by performing the task in person and demonstrating the correct actions to the learning agent. Most imitation learning or learning from demonstration algorithms assume this type of guidance~\cite{schaal1999imitation,argall2009survey,osa2018algorithmic}. Nevertheless, in many cases it is impractical to use human policy as guidance because some of these tasks are too challenging for even humans to perform well. Additionally, IL algorithms may require a large amount of high-quality demonstration data, whereas collecting human behavioral data could be expensive and subject to error. 

One possible solution is to leverage other types of human guidance. The types of human guidance we discuss here can be seen as feedback. The intuition is that even for tasks that are too challenging for humans, they still could provide feedback regarding the performance and guide the agent in that respect. These types of guidance could be less expensive than policy demonstration, or can be collected in parallel with policy demonstration to provide additional information. 

This survey aims at providing a high-level overview of recent research efforts that primarily rely on these types of human guidance other than conventional, step-by-step action demonstrations to solve complex deep imitation and reinforcement learning tasks. The types of guidance we review include human evaluative feedback, human preference, high-level goals, human attention, and state sequences without actions. The corresponding approaches we discuss vary with regards to the trade-off between the amount of information provided to the agent and the amount of human effort required. All of these approaches have shown promising results in one or more challenging deep reinforcement learning tasks, such as Atari video games~\cite{bellemare2012arcade} and simulated robotic control~\cite{todorov2012mujoco}.

\section{Background}

\begin{figure*}[ht!]
\centering
\subfloat[Standard imitation learning]{\includegraphics[width=0.25\textwidth]{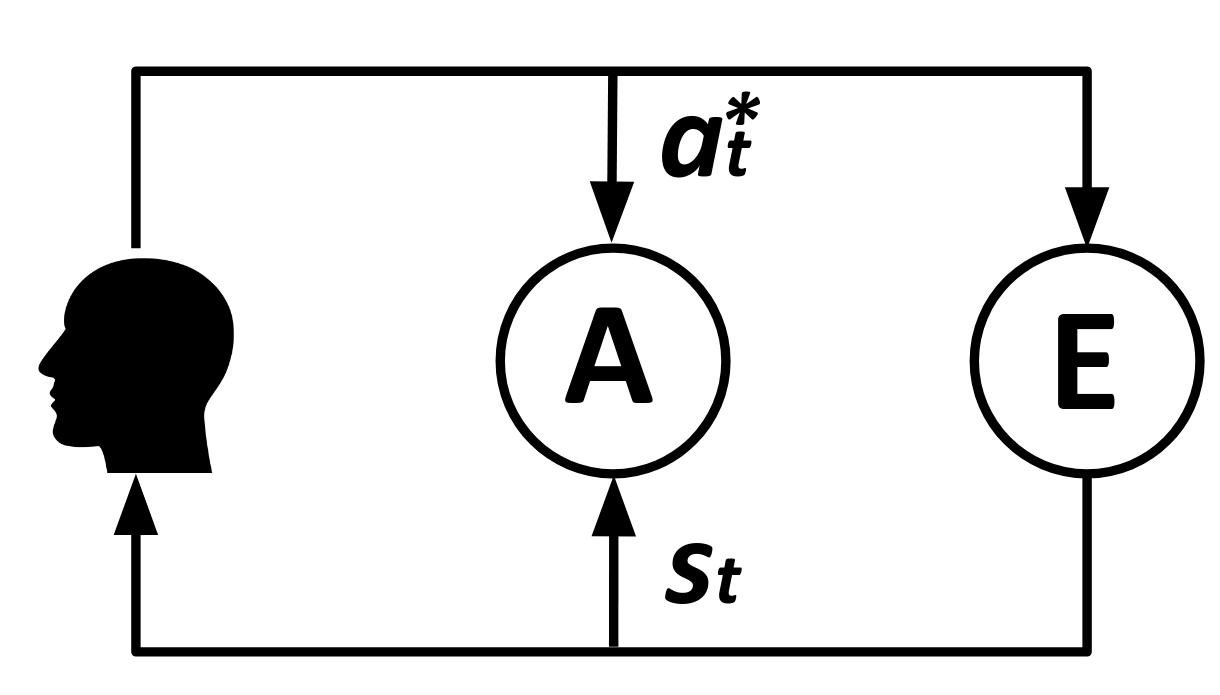}\label{fig:il}}
\subfloat[Evaluative feedback]{\includegraphics[width=0.25\textwidth]{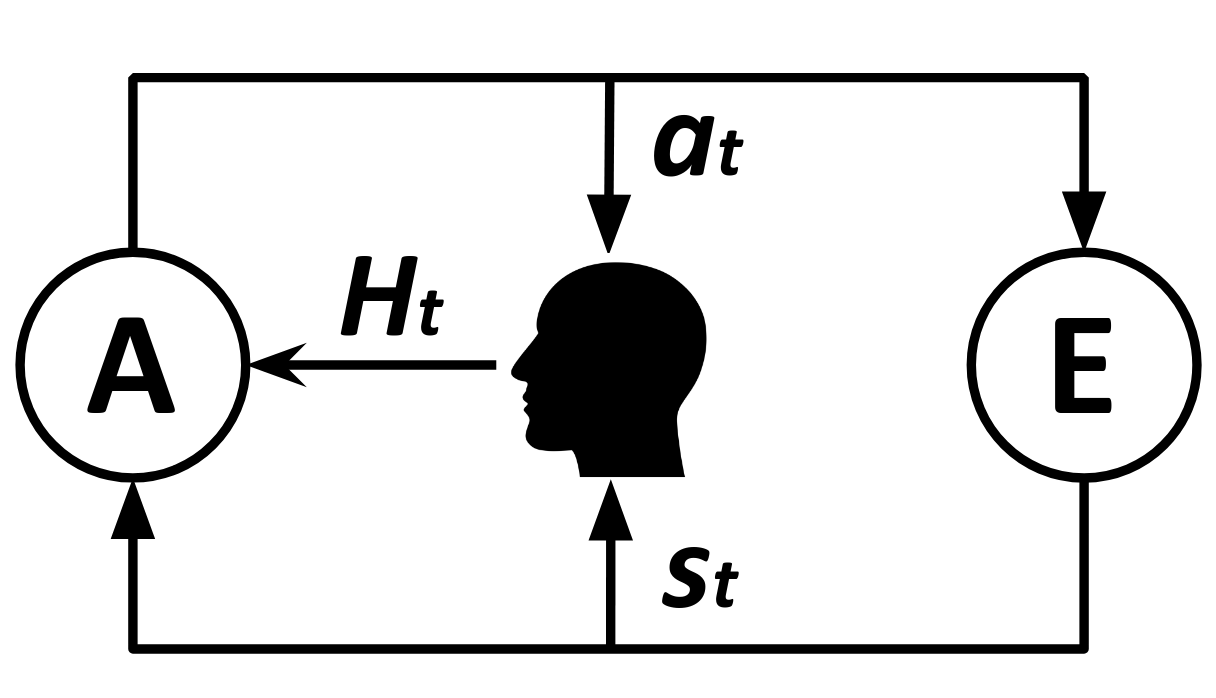}\label{fig:fb}}
\subfloat[Imitation from observation]{\includegraphics[width=0.25\textwidth]{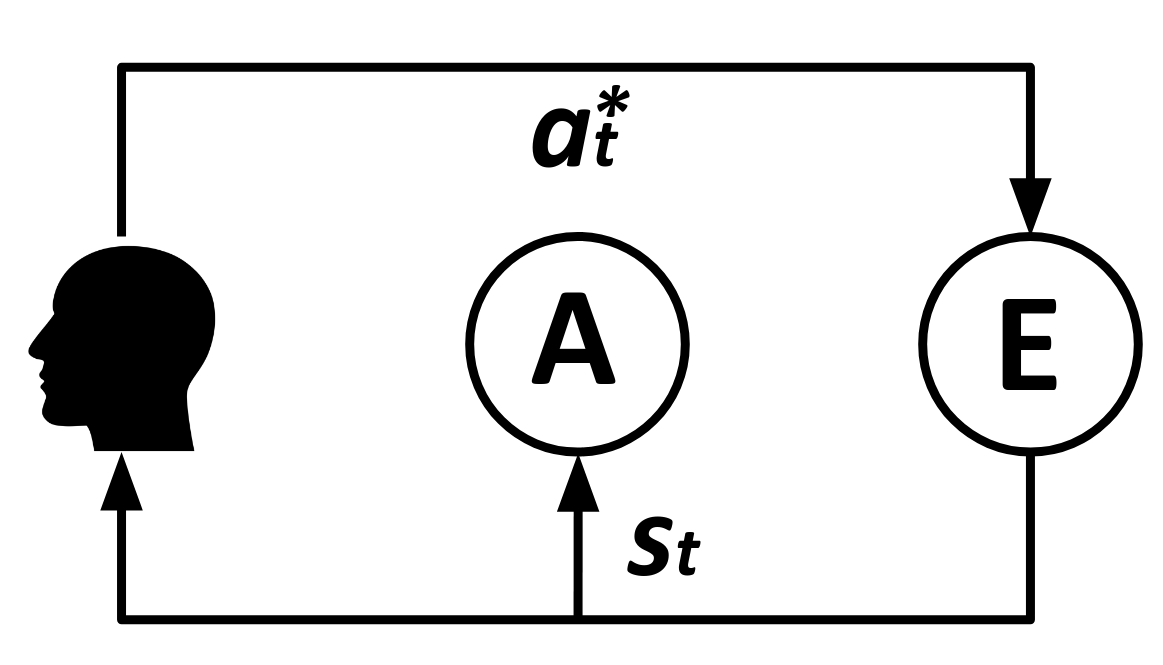}\label{fig:ifo}}
\subfloat[Learning attention from human]{\includegraphics[width=0.25\textwidth]{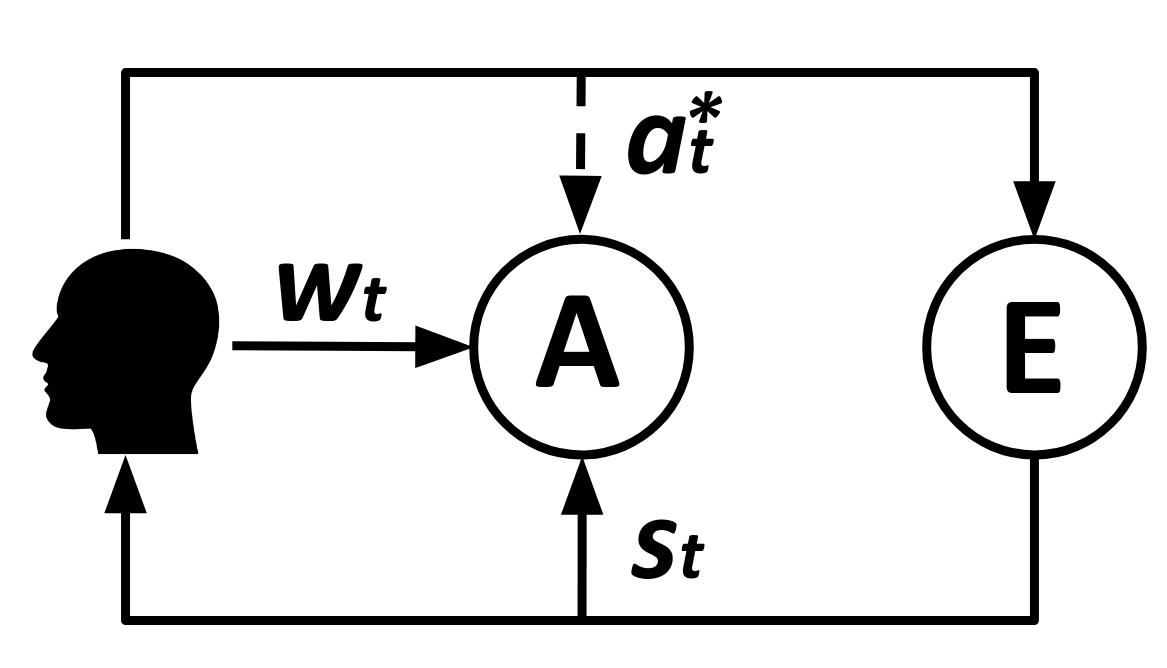}\label{fig:att}}
\caption{Human-agent-environment interaction diagrams of different approaches discussed in this paper. These diagrams illustrate how different types of human guidance data are collected, including information required by the human trainer and the guidance provided to the agent. Note that the learning process of the agent is not included in these diagrams. \textbf{A}: learning agent; \textbf{E}: environment; Arrow: information flow direction; Dashed arrow: optional information flow. In (a) standard imitation learning, the human trainer observes state information $s_t$ and demonstrates action $a^*_t$ to the agent; the agent stores this data to be used in learning later. In (b) learning from evaluative feedback, the human trainer watches the agent performing the task, and provides instant feedback $H_t$ on agent decision $a_t$ in state $s_t$. (c) Imitation from observation is similar to standard imitation learning except that the agent does not have access to human demonstrated action. (d) Learning attention from human requires the trainer to provide attention map $w_t$ to the learning agent. }
\label{fig:approach}
\end{figure*}

\begin{figure}[ht!]
\centering
\includegraphics[width=0.325\textwidth]{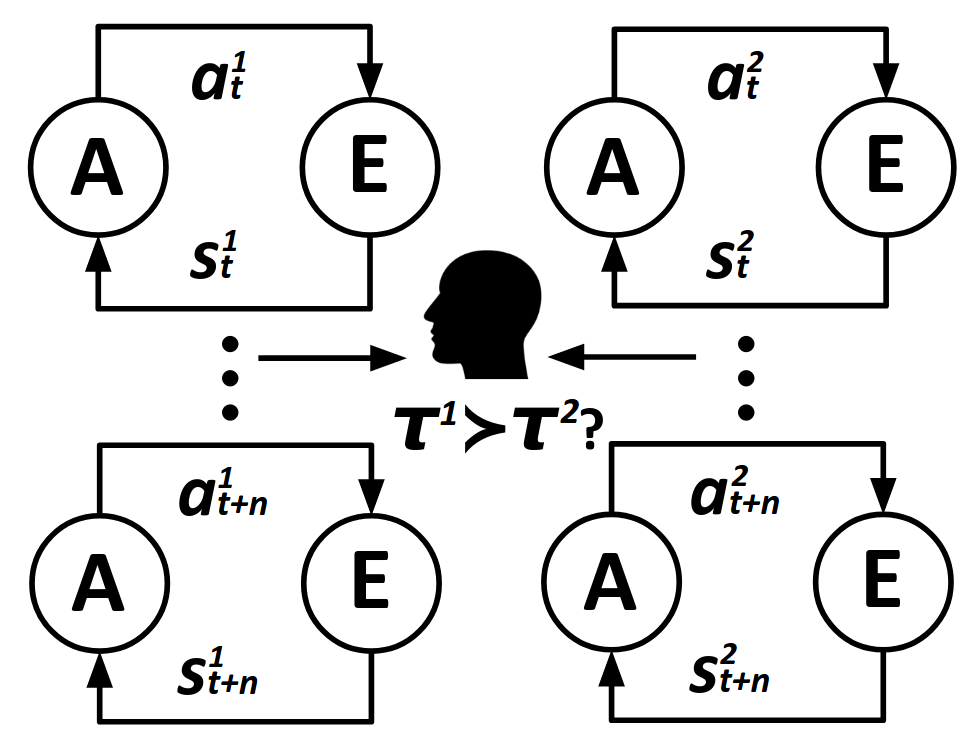}
\caption{Learning from human preference. The human trainer watches two behaviors generated by the learning agent simultaneously, an decides which behavior is more preferable. $\tau^1 \succ \tau^2$ denotes that the trainer prefers behavior trajectory $\tau^1$ over $\tau^2$.}
\label{fig:pref}
\end{figure}

\begin{figure}[ht!]
\centering
\includegraphics[width=0.29\textwidth]{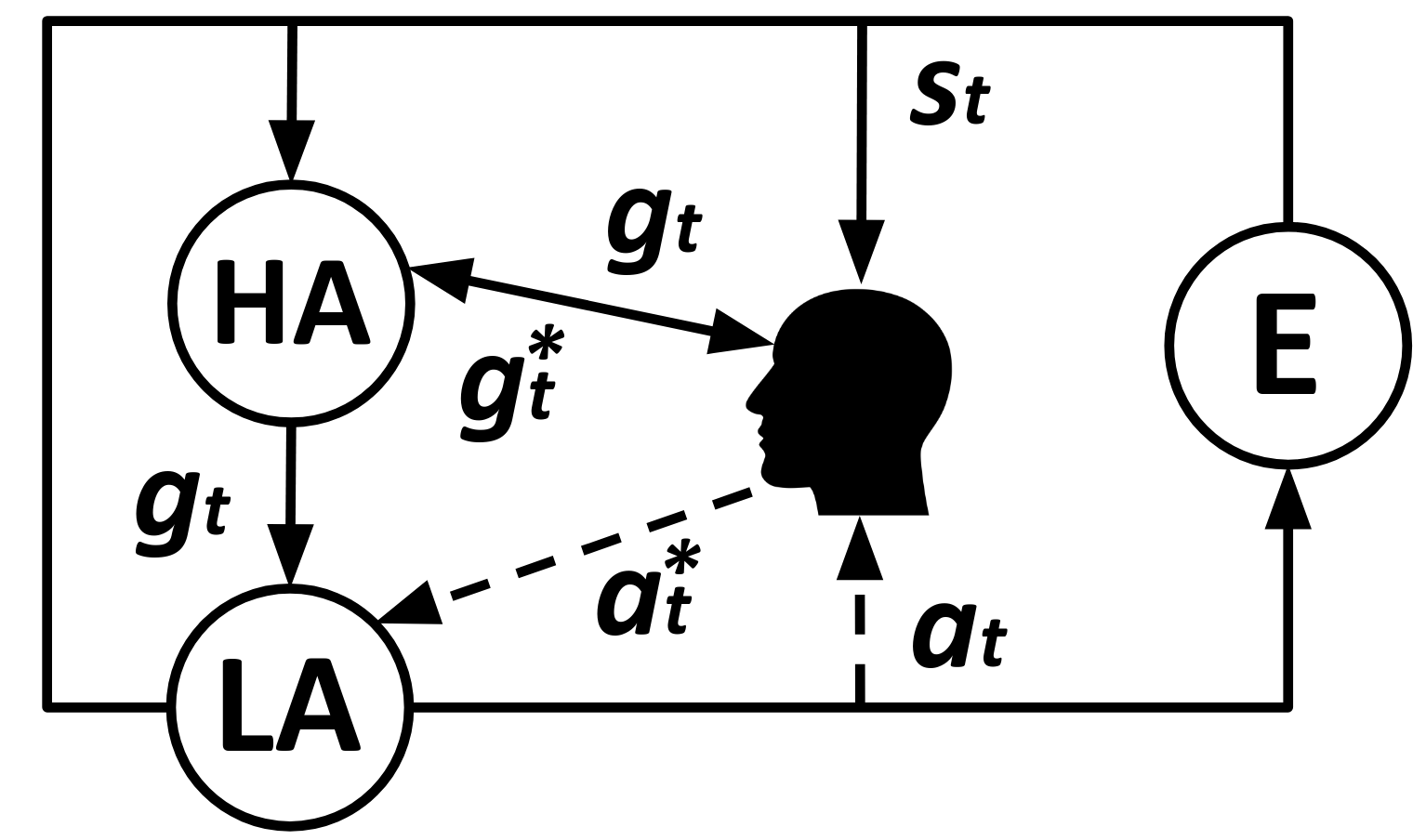}
\caption{Hierarchical imitation. \textbf{HA}: high-level agent; \textbf{LA}: low-level agent. The high-level agent chooses a high-level goal $g_t$ for state $s_t$. The low-level agent then chooses an action $a_t$ based on $g_t$ and $s_t$. The primary guidance that the trainer provides in this framework is the correct high-level goal $g^*_t$.}
\label{fig:hi}
\end{figure}

A standard reinforcement learning problem is formalized as a Markov decision process (MDP), defined as a tuple $\langle\S, \A, \T, \R, \gamma\rangle$~\cite{sutton2018reinforcement}. $\S$ is a set of environment states which encode relevant information for an agent's decision. $\A$ is a set of agent actions. $\T$ is the state transition function which describes $p(s'|s,a)$, i.e., the probability of entering state $s'$ when an agent takes action $a$ in state $s$. $\R$ is a reward function. $r(s,a,s')$ denotes the scalar reward agent received on transition from $s$ to $s'$ under action $a$. $\gamma \in \lbrack 0,1 \rbrack$ is a discount factor that indicates how much the agent value an immediate reward compared to a future reward. 

$\pi: \S \times \A \mapsto \lbrack 0,1 \rbrack$ is a policy which specifies the probability distribution of selecting actions in a given state. The goal for a learning agent is to find an optimal policy $\pi^*$ that maximizes the expected cumulative reward. One could optimize $\pi$ directly, while alternatively many of the algorithms are based on value function estimation, i.e., estimating the state value function $V^{\pi}(s)$ or the action-value function $Q^{\pi}(s,a)$. Nowadays, deep neural networks are often used as function approximators to estimate and optimize $\pi$, $V$, and $Q$.

The standard imitation learning setting (Fig.~\ref{fig:il}) can be formulated as MDP\textbackslash$R$, i.e. there is no reward function $\R$ available. Instead, a learning agent records human demonstration in the format of state-action pairs $\{(s_t,a^*_t), (s_{t+1},a^*_{t+1}) \dots\}$ at each timestep. Then the agent can learn to directly mimic the human demonstrated policy using supervised learning, an approach known as behavioral cloning~\cite{bain1999framework}. Or it can estimate the human reward function from demonstrated policy as an intermediate step followed by standard RL to learn a policy using this reward function which is known as inverse reinforcement learning (IRL)~\cite{abbeel2004apprenticeship}. These two approaches are the major learning frameworks used in imitation learning and comprehensive reviews of these approaches can be found in~\cite{hussein2017imitation,osa2018algorithmic}. Both approaches assume that $(s_t,a_t)$ pairs are the only learning signal to the agent, and both $s_t$ and $a_t$ are available. In contrast, the learning frameworks we are about to discuss utilize different learning signals.

\section{Learning from Evaluative Feedback}
The first type of human guidance we consider is real-time evaluative feedback, which is given by a human trainer while watching the agent performing the task (Fig.~\ref{fig:fb}). The agent adjusts its policy online based on the feedback received. The simplest form of evaluative feedback is a scalar value indicating how desirable an observed action is. Compared to policy demonstration, this approach greatly reduces the needed human effort. It does not necessarily require the human trainer to be an expert at \emph{performing} the task--it only requires the trainer to accurately \emph{judge} agent behaviors. 

One of the main challenges in this approach is to interpret human feedback correctly since such interpretation determines how the feedback is used to improve the policy in the MDP framework. Here we compare various methods that assume different interpretations of human feedback.

\subsection{Policy Shaping} 
A policy shaping approach, e.g., the Advise framework, interprets human feedback as direct policy labels~\cite{griffith2013policy,cederborg2015policy}. The feedback could simply be ``right" or ``wrong", indicating whether the observed action is optimal or not. If this feedback is consistent, the agent could simply change its policy to always choose the optimal action confirmed by the human, and eliminate actions that are wrong. ~\cite{griffith2013policy} assumes such feedback could be inconsistent. Assuming human trainers provide the correct feedback with probability $\mathcal{C}$. The decision policy thus can be represented using a binomial distribution $\pi(s,a) = \frac{\mathcal{C}^{\Delta s,a}}{\mathcal{C}^{\Delta s,a} + (1-\mathcal{C})^{\Delta s,a}}$ where $\Delta s,a$ is the difference between the number
of “right” and “wrong” labels for action $a$ in state $s$. Explicitly modeling this inconsistency in human feedback allowed Advise to outperform other methods using synthetic feedback~\cite{griffith2013policy} and was validated on human trainers~\cite{cederborg2015policy}. 

\subsection{Reward Shaping} In contrast to policy shaping, reward shaping interprets human feedback as the value function or a human-specified reward function~\cite{isbell2001social,tenorio2010dynamic,pilarski2011online}. Note that this reward function may completely replace the reward function provided by the environment. The TAMER (Training an Agent Manually via Evaluative Reinforcement) framework adopts the reward function interpretation~\cite{knox2009interactively} and has been recently extended to deep TAMER~\cite{warnell2018deep}. 

TAMER assumes that the human trainer has an internal reward function $H: \S \times \A \mapsto \{-1,0,+1\}$ that maps an observed agent action in a state to a negative, neutral, or positive feedback. Deep TAMER uses a deep network to represent and learn an estimate of $H$, $\hat{H}$, via supervised learning. The agent chooses the action $\pi(s) = \max_a \hat{H}(s,a)$. 
One notable result from~\cite{warnell2018deep} is that they asked human trainers to perform the task, and compared a deep TAMER agent's final performance with human trainers'. They found that the agent was able to outperform humans. The result confirms that the feedback-based approach does not require the humans to be experts at performing the task and the learning agent's performance is not capped by the trainers' expertise. 

A natural extension is to combine human reward signal with environment reward signal, an approach named TAMER+RL~\cite{knox2010combining,knox2012reinforcement}. Recently deep TAMER has been combined with a Deep Q-Network~\cite{mnih2015human} in the DQN-TAMER framework~\cite{arakawa2018dqn}. A DQN agent and a TAMER agent are trained in parallel, and the final decision policy is a weighted average of the policies from both agents. Effective methods to combine two sources of reward are yet to be explored although several possibilities have been proposed~\cite{knox2010combining}.


\subsection{Intervention} A slightly different but related approach is Human Intervention RL~\cite{saunders2018trial}, in which a human supervises the training process of an RL agent and blocks catastrophic actions. The blocked state-action pairs can be used to train a binary classifier to detect and block unsafe actions~\cite{saunders2018trial}. Since blocking can be viewed as strong negative feedback, the intervention method can be seen as TAMER+RL without positive feedback.

\subsection{Policy-dependent Feedback} TAMER interprets human feedback as a reward function that is independent of the agent's current policy. However, COACH (Convergent Actor-Critic by Humans)~\cite{macglashan2017interactive} assumes that such feedback is policy-dependent, and should be better interpreted as the advantage function that specifies how much better or worse when deviating from the agent's current policy. Note that the advantage function gradually decreases to zero or negative as $\pi$ improves. This interpretation better captures several phenomena observed in human feedback such as diminishing returns. The human feedback can then be used to replace the advantage function in calculating the policy gradient using the actor-critic algorithm~\cite{macglashan2017interactive}. 

The algorithms above interpret human feedback differently, and result in different reinforcement learning objectives and update rules. Using synthetic feedback, \citeauthor{macglashan2017interactive} showed that the convergence of these algorithms depends critically on whether the actual feedback matches the assumed one~\shortcite{macglashan2017interactive}. The nature of the feedback could potentially vary across tasks and trainers~\cite{loftin2016learning}, and can be altered by instruction given to the trainers~\cite{cederborg2015policy}. Therefore these factors need to be carefully controlled in practice. A potential further direction is to design a method that is either robust to all types of feedback, or can infer the human feedback type using a probabilistic model and adapt to the trainer's strategy online~\cite{loftin2016learning}.


\section{Learning from Human Preference}
The second type of guidance we consider is human preference. Many RL tasks are difficult for humans to provide demonstrations or evaluative feedback. These could be control tasks with many degrees of freedom with non-human morphology, e.g., MuJoCo~\cite{todorov2012mujoco}, it is therefore harder for humans to control or to tell whether a particular state-action pair is good or not. It is more natural for the agent to query human trainers for their \emph{preferences} or \emph{rankings} over a set of behaviors. This feedback can be provided over a set of state or action sequences; however, it is much less demanding if it is over trajectories, as the trainer can directly evaluate the outcome of the trajectories. As a result, as shown in Fig.~\ref{fig:pref}, we consider preferences over trajectory segments, or sequences of state-action pairs: $\tau = ((s_0,a_0),(s_1,a_1),\dots)$. 

Previous works have used preferences to directly learn policies \cite{wilson2012bayesian,busa2013preference}, learn a preference model \cite{furnkranz2012preference}, or learn a reward function \cite{wirth2016model,akrour2014programming}.
A survey on these topics is provided by~\citeauthor{wirth2017survey} \shortcite{wirth2017survey}. 

More recent works have extended previous preference-based RL methods to be compatible with deep RL~\cite{christiano2017deep,ibarz2018reward}. A pair of agent trajectories of 1-2 seconds long are simultaneously presented to human trainers to query for their preference. The goal is to learn the latent human reward function $r(s,a)$, as in IRL, but from human preference instead of a human policy. Intuitively, the probability of a human preferring a segment could depend on the total reward summed over the trajectory~\cite{christiano2017deep}. A model can be trained to find the reward function that minimizes the cross-entropy loss between model's prediction and human preference. Since the targets to be evaluated are trajectories instead of state-action pairs, the feedback can be very sparse. This approach drastically reduces human effort. The amount of human feedback required can be as little as 1\% of the total number of agent actions~\cite{christiano2017deep}. \citeauthor{ibarz2018reward} further extended this approach by including human demonstration to pre-train the agent, and including demonstration trajectories in the preference learning, assuming human trajectories are always more preferable than agent trajectories~\shortcite{ibarz2018reward}. 

An important component in this framework is query selection, in which the agent needs to decide which trajectories to query for preference. \citeauthor{christiano2017deep} simply select trajectories such that an ensemble of learning models have the largest variance in predicting their human preferences~\shortcite{christiano2017deep}. Ideally, the query should maximize the expected information gain from an active learning perspective, an important research challenge that is closely related to preference elicitation~\cite{zintgraf2018ordered}.


\section{Hierarchical Imitation}
Many reinforcement learning tasks are hierarchically structured, meaning that they can be decomposed and solved using a divide-and-conquer approach. It is possible to ask human trainers to only provide high-level feedback on these tasks. Similar to preference, this type of feedback also targets trajectory segments, but is provided as option choice~\cite{sutton1999between}, i.e., choice of high-level goal in a given state. Due to the task's hierarchical structure, the behavior trajectory can be naturally segmented into these options, instead of arbitrarily segmented in the preference framework.

\citeauthor{le2018hierarchical} \shortcite{le2018hierarchical} has proposed the hierarchical guidance framework that assumes a two-level hierarchy, in which a high-level agent learns to choose a goal $g$ given a state, while low-level agents learn to execute a sub-policy (option) to accomplish the chosen goal (Fig.~\ref{fig:hi}). Note that an action to terminate the current option needs to be added to the low-level agent's action space, and this termination action can be demonstrated by a human and learned by the agent. Human trainers are asked to provide three types of feedback: 1) a positive signal if the high-level goal $g_t$ and low-level sub-policy $a_t$ are both correct. 2) the correct high-level goal $g^*_t$ for state $s_t$ if the chosen $g_t$ is incorrect. 3) the correct low-level action $a^*_t$ if the high-level goal was chosen correctly but the low-level sub-policy is incorrect. At each level, the learning task becomes a typical imitation learning problem, therefore conventional IL algorithms such as behavioral cloning and DAgger~\cite{ross2011reduction} can be applied. Perhaps the most exciting result comes from a hybrid of hierarchical imitation learning and RL, in which a DAgger agent learns to choose a high-level goal from human guidance, and a Q-learning agent learns low-level sub-policies. This approach was shown to be substantially more sample efficient than conventional imitation learning on a very difficult Atari game called Montezuma's revenge~\cite{le2018hierarchical} . 

\citeauthor{andreas2017modular} take a similar hierarchical approach but only require humans to provide policy sketches that are high-level symbolic subtask labels~\shortcite{andreas2017modular}. The policy of each subtask is learned by the RL agent on its own and no longer requires human demonstration. As mentioned earlier, it is natural to extend hierarchical imitation to incorporate human preferences over the outcome of options, instead of asking humans to provide the correct option labels, as done in~\citeauthor{pinsler2018sample}~\shortcite{pinsler2018sample}.

\section{Imitation from Observation}
Imitation from observation (IfO) \cite{torabi2019recent} is the problem of learning directly by observing a trainer performing the task. The learning agent only has access to visual data of the trainer; therefore, no action data is available (Fig.~\ref{fig:ifo}). The ultimate goal in this framework is to enable agents to utilize existing, rich amount of demonstration data that do not have action labels, such as online videos of humans performing various tasks. There are two general challenges in addressing this problem: (1) perception, and (2) control.

Existing resources are not necessarily generated in ideal conditions; therefore, the agent should be able to deal with the challenges in perception. For instance, the demonstrations could have different viewpoints. One approach developed to address this issue learns a context translation model to translate an observation to predict the observation in the target context \cite{liu2018imitation}. Another approach uses a classifier to distinguish between the data that comes from different viewpoints and attempts to maximize the domain confusion in an adversarial setting during the training. Consequently, the extracted features would be invariant with respect to the viewpoint \cite{stadie2017third}. One other challenge in perception is embodiment mismatch where the imitator and the trainer do not share the same embodiment. One method developed to approach this problem learns a correspondence between the embodiments using autoencoders in a supervised fashion \cite{gupta2017learning}. Another method learns the correspondence in an unsupervised fashion with a small number of human supervision \cite{sermanet2018time}.

The second challenge in IfO is control. This component concerns learning the task from proprioceptive states of the trainer $(s_0, s_1,\dots)$ instead of visual data. The existing IfO control algorithms can be categorized in two broad classes: model-based and model-free.

Model-based algorithms learn a dynamics model during the imitation process. One approach is to learn an inverse dynamics model which estimates the taken action $a$, given a state transitions $(s, s')$. Therefore, this model can be used to infer the missing action labels of the expert. Then, the inferred actions can be executed to reproduce the trainers states \cite{nair2017combining}. As an alternative, after inferring the actions, a mapping from states to the actions can be learned and used to improve the learned model and consequently the policy \cite{torabi2018behavioral}. If a sparse reward function is available, this algorithm can be combined with an RL algorithm to improve the performance \cite{guo2019hybrid}. Another type of dynamics model is a forward dynamics model which predicts the next state $s'$, given the current state and action $(s,a)$. A combination of policy with this type of dynamics model can be used to predict the next state from a given current state and then train the imitation policy to make the prediction more accurate \cite{edwards2018imitating}.

On the other hand, model-free algorithms imitate the task without explicitly learning any dynamics model. A class of this type of algorithms imitate by designing a reward function as the Euclidean distance of the states of the imitator from that of the trainer at each timestep, and then solving an RL problem \cite{sermanet2018time,gupta2017learning,dwibedi2018learning}. Another class of approaches, inspired by generative adversarial imitation learning \cite{ho2016generative}, imitate the tasks in an adversarial fashion. A method of this type uses GANs~\cite{goodfellow2014generative} to bring the state distribution of the imitator closer to that of the trainer \cite{merel2017learning}. However, since equivalence in {\em single state} distribution does not mean equivalence in policy, getting the {\em state-transition} distributions closer together may result in better performance \cite{torabi2018generative,torabi2019adversarial,torabi2019imitation}.

\section{Learning Attention from Humans}
During the human demonstration or evaluation process, there are other helpful signals for a learning agent from a social learning perspective. For example, human voice~\cite{tenorio2010dynamic} and facial expression~\cite{arakawa2018dqn} can be used as additional sources of evaluative feedback. For visual learning tasks, a particularly useful information source from humans is the gaze position, which can be collected in parallel with demonstration. Researchers have collected human gaze and policy data for daily cooking \cite{li2018eye}, Atari game playing \cite{zhang2019atari}, and outdoor driving \cite{palazzi2018predicting,kim2018textual}. Since human eyes have limited resolution except for the center fovea, humans learn to move their eyes to the correct place at the right time to process urgent state information. The agent could learn to extract useful state features by learning to attend from human. Recent works have trained convolutional neural networks to predict the probability distribution of human gaze given a raw image and were able to generate a gaze heatmap in the pixel space~\cite{li2018eye,palazzi2018predicting,kim2018textual,zhang2019atari}. This heatmap serve as a strong feature extractor that is particularly helpful for tasks with a high-dimensional state space.

Existing works have proposed to use a learned gaze prediction model to guide the process of learning a human policy \cite{li2018eye,zhang2018agil,kim2018textual}. Intuitively, knowing where humans would look provides useful information on what action they will take. Therefore the predicted gaze heatmap helps select important features in the given state. Doing so results in higher accuracy in imitating human actions~\cite{li2018eye,zhang2018agil,kim2018textual} and incorporating learned gaze model into behavioral cloning leads to an average performance increase of 198\% in Atari games~\cite{zhang2018agil}. 

For future work, attention learning is closely related to hierarchical imitation, since gaze is a good indicator of the current high-level behavioral goal which might help an imitator to infer the correct low-level action. Furthermore, gaze data can be collected in parallel with evaluation. This data might reveal more information to the learning agent to explain why the human gives a particular evaluative feedback.


\section{Conclusion and Future Directions}
This survey aims at providing a literature review for progress in leveraging different types of human guidance to solve deep reinforcement learning tasks. Here we briefly discuss several promising future research directions. 

\subsection{Shared Datasets and Reproducibility} In general, researchers collect their own human guidance data. However, this type of data is often expensive to collect. An effort that could greatly facilitate research in this field is to create publicly available benchmark datasets. Collecting dataset would be difficult for interactive learning methods; but for other approaches, data can be collected in advance and shared. Another concern is reproducibility in RL~\cite{henderson2018deep}. When collecting human guidance data, factors such as individual expertise, experimental setup, data collection tools, dataset size, and experimenter bias could introduce large variances in final performance. Therefore, evaluating algorithms using a standard dataset could save the effort and assure a fair comparison between algorithms. But it would not allow for feedback that is dependent on the changing policy as it is being learned.

\subsection{Understanding Human Trainers} Leveraging human guidance to train an agent naturally follows a teacher-student paradigm. Much effort has been spent on making the student more intelligent. However, understanding the behavior of human teachers is equally important. \citeauthor{thomaz2008teachable} pioneered the effort in understanding human behavior in teaching learning agents~\shortcite{thomaz2008teachable}. As RL agents become more powerful and attempt to solve more complex tasks, the human teachers' guiding behaviors could become more complicated and require further study. 

Studying the behaviors, especially the limitations of human teachers, allows one to design a teaching environment that is more effective and produces more useful guidance data. \citeauthor{amir2016interactive}~\shortcite{amir2016interactive} studied human attention limits while monitoring the learning process of an agent and proposed an algorithm for the human and the agent to jointly identify states where feedback is most needed to reduce human monitoring cost. \citeauthor{ho2016showing} showed the differences in behavior when a human trainer is intentionally teaching (showing) versus merely doing the task. They found that humans modify their policies to reveal the goal to the agent when in the showing mode but not in doing mode. They further showed that imitation learning algorithm can benefit substantially more from the data collected in the showing mode~\cite{ho2016showing}.

Understanding the nature of human guidance allows algorithms to learn more effectively. We have already seen the debate on how to interpret human evaluative feedback in complex tasks. A helpful way to resolve this debate is to conduct human studies with diverse subject pools to investigate whether real-life human feedback satisfies their algorithmic assumptions and what factors affect the human feedback strategy~\cite{cederborg2015policy,loftin2016learning,macglashan2017interactive}.  

\subsection{A Unified Lifelong Learning Paradigm} The learning frameworks discussed in this paper are often inspired by real-life biological learning scenarios that correspond to different learning stages and strategies in lifelong learning. Imitation and reinforcement learning correspond to learning completely by imitating others and learning completely through self-generated experience, where the former may be used more often in early stages of learning and the latter could be more useful in late stages. The other learning strategies discussed fill in the gap between those two extremes. In reality, an animal is likely to utilize all possible intermediate learning signals to learn to perform a challenging task to gain reward. We have compared these approaches within an imitation and reinforcement learning framework. Under this framework, it is possible to develop a unified learning paradigm that accepts multiple types of human guidance, as explored by \citeauthor{abel2017agent}~\shortcite{abel2017agent}.

\section*{Acknowledgements}
Part of this work is supported by NIH grant EY 05729. A portion of this work has taken place in the Learning Agents Research Group (LARG) at UT Austin.  LARG research is supported in part by NSF (IIS-1637736, IIS-1651089, IIS-1724157), ONR (N00014-18-2243), FLI (RFP2-000), ARL, DARPA, Intel, Raytheon, and Lockheed Martin.  Peter Stone serves on the Board of Directors of Cogitai, Inc. The terms of this arrangement have been reviewed and approved by the UT Austin in accordance with its policy on objectivity in research.

\small{
\bibliographystyle{named}
\bibliography{ref.bib}

\begin{thebibliography}{}

\bibitem[\protect\citeauthoryear{Abbeel and
  Ng}{2004}]{abbeel2004apprenticeship}
Pieter Abbeel and Andrew~Y Ng.
\newblock Apprenticeship learning via inverse reinforcement learning.
\newblock In {\em Proceedings of the twenty-first international conference on
  Machine learning}, page~1. ACM, 2004.

\bibitem[\protect\citeauthoryear{Abel \bgroup \em et al.\egroup
  }{2017}]{abel2017agent}
David Abel, John Salvatier, Andreas Stuhlm{\"u}ller, and Owain Evans.
\newblock Agent-agnostic human-in-the-loop reinforcement learning.
\newblock {\em NeurIPS Workshop on the Future of Interactive Learning
  Machines}, 2017.

\bibitem[\protect\citeauthoryear{Akrour \bgroup \em et al.\egroup
  }{2014}]{akrour2014programming}
Riad Akrour, Marc Schoenauer, Mich{\`e}le Sebag, and Jean-Christophe Souplet.
\newblock Programming by feedback.
\newblock In {\em International Conference on Machine Learning}, volume~32,
  pages 1503--1511. JMLR. org, 2014.

\bibitem[\protect\citeauthoryear{Amir \bgroup \em et al.\egroup
  }{2016}]{amir2016interactive}
Ofra Amir, Ece Kamar, Andrey Kolobov, and Barbara~J Grosz.
\newblock Interactive teaching strategies for agent training.
\newblock In {\em Proceedings of the Twenty-Fifth International Joint
  Conference on Artificial Intelligence}, pages 804--811. AAAI Press, 2016.

\bibitem[\protect\citeauthoryear{Andreas \bgroup \em et al.\egroup
  }{2017}]{andreas2017modular}
Jacob Andreas, Dan Klein, and Sergey Levine.
\newblock Modular multitask reinforcement learning with policy sketches.
\newblock In {\em Proceedings of the 34th International Conference on Machine
  Learning-Volume 70}, pages 166--175. JMLR. org, 2017.

\bibitem[\protect\citeauthoryear{Arakawa \bgroup \em et al.\egroup
  }{2018}]{arakawa2018dqn}
Riku Arakawa, Sosuke Kobayashi, Yuya Unno, Yuta Tsuboi, and Shin-ichi Maeda.
\newblock Dqn-tamer: Human-in-the-loop reinforcement learning with intractable
  feedback.
\newblock {\em arXiv preprint arXiv:1810.11748}, 2018.

\bibitem[\protect\citeauthoryear{Argall \bgroup \em et al.\egroup
  }{2009}]{argall2009survey}
Brenna~D Argall, Sonia Chernova, Manuela Veloso, and Brett Browning.
\newblock A survey of robot learning from demonstration.
\newblock {\em Robotics and autonomous systems}, 57(5):469--483, 2009.

\bibitem[\protect\citeauthoryear{Bain and Sommut}{1999}]{bain1999framework}
Michael Bain and Claude Sommut.
\newblock A framework for behavioural claning.
\newblock {\em Machine intelligence}, 15(15):103, 1999.

\bibitem[\protect\citeauthoryear{Bellemare \bgroup \em et al.\egroup
  }{2012}]{bellemare2012arcade}
Marc~G Bellemare, Yavar Naddaf, Joel Veness, and Michael Bowling.
\newblock The arcade learning environment: An evaluation platform for general
  agents.
\newblock {\em Journal of Artificial Intelligence Research}, 2012.

\bibitem[\protect\citeauthoryear{Busa-Fekete \bgroup \em et al.\egroup
  }{2013}]{busa2013preference}
R{\'o}bert Busa-Fekete, Bal{\'a}zs Sz{\"o}r{\'e}nyi, Paul Weng, Weiwei Cheng,
  and Eyke H{\"u}llermeier.
\newblock Preference-based evolutionary direct policy search.
\newblock In {\em ICRA Workshop on Autonomous Learning}, 2013.

\bibitem[\protect\citeauthoryear{Cederborg \bgroup \em et al.\egroup
  }{2015}]{cederborg2015policy}
Thomas Cederborg, Ishaan Grover, Charles~L Isbell, and Andrea~L Thomaz.
\newblock Policy shaping with human teachers.
\newblock In {\em Twenty-Fourth International Joint Conference on Artificial
  Intelligence}, 2015.

\bibitem[\protect\citeauthoryear{Christiano \bgroup \em et al.\egroup
  }{2017}]{christiano2017deep}
Paul~F Christiano, Jan Leike, Tom Brown, Miljan Martic, Shane Legg, and Dario
  Amodei.
\newblock Deep reinforcement learning from human preferences.
\newblock In {\em Advances in Neural Information Processing Systems}, pages
  4299--4307, 2017.

\bibitem[\protect\citeauthoryear{Dwibedi \bgroup \em et al.\egroup
  }{2018}]{dwibedi2018learning}
Debidatta Dwibedi, Jonathan Tompson, Corey Lynch, and Pierre Sermanet.
\newblock Learning actionable representations from visual observations.
\newblock In {\em 2018 IEEE/RSJ International Conference on Intelligent Robots
  and Systems (IROS)}, pages 1577--1584. IEEE, 2018.

\bibitem[\protect\citeauthoryear{Edwards \bgroup \em et al.\egroup
  }{2018}]{edwards2018imitating}
Ashley~D Edwards, Himanshu Sahni, Yannick Schroeker, and Charles~L Isbell.
\newblock Imitating latent policies from observation.
\newblock {\em arXiv preprint arXiv:1805.07914}, 2018.

\bibitem[\protect\citeauthoryear{F{\"u}rnkranz \bgroup \em et al.\egroup
  }{2012}]{furnkranz2012preference}
Johannes F{\"u}rnkranz, Eyke H{\"u}llermeier, Weiwei Cheng, and Sang-Hyeun
  Park.
\newblock Preference-based reinforcement learning: a formal framework and a
  policy iteration algorithm.
\newblock {\em Machine learning}, 89(1-2):123--156, 2012.

\bibitem[\protect\citeauthoryear{Goodfellow \bgroup \em et al.\egroup
  }{2014}]{goodfellow2014generative}
Ian Goodfellow, Jean Pouget-Abadie, Mehdi Mirza, Bing Xu, David Warde-Farley,
  Sherjil Ozair, Aaron Courville, and Yoshua Bengio.
\newblock Generative adversarial nets.
\newblock In {\em Advances in neural information processing systems}, pages
  2672--2680, 2014.

\bibitem[\protect\citeauthoryear{Griffith \bgroup \em et al.\egroup
  }{2013}]{griffith2013policy}
Shane Griffith, Kaushik Subramanian, Jonathan Scholz, Charles~L Isbell, and
  Andrea~L Thomaz.
\newblock Policy shaping: Integrating human feedback with reinforcement
  learning.
\newblock In {\em Advances in neural information processing systems}, pages
  2625--2633, 2013.

\bibitem[\protect\citeauthoryear{Guo \bgroup \em et al.\egroup
  }{2019}]{guo2019hybrid}
Xiaoxiao Guo, Shiyu Chang, Mo~Yu, Gerald Tesauro, and Murray Campbell.
\newblock Hybrid reinforcement learning with expert state sequences.
\newblock {\em Association for the Advancement of Artificial Intelligence},
  2019.

\bibitem[\protect\citeauthoryear{Gupta \bgroup \em et al.\egroup
  }{2018}]{gupta2017learning}
Abhishek Gupta, Coline Devin, YuXuan Liu, Pieter Abbeel, and Sergey Levine.
\newblock Learning invariant feature spaces to transfer skills with
  reinforcement learning.
\newblock In {\em International Conference on Learning Representations}, 2018.

\bibitem[\protect\citeauthoryear{Henderson \bgroup \em et al.\egroup
  }{2018}]{henderson2018deep}
Peter Henderson, Riashat Islam, Philip Bachman, Joelle Pineau, Doina Precup,
  and David Meger.
\newblock Deep reinforcement learning that matters.
\newblock In {\em Thirty-Second AAAI Conference on Artificial Intelligence},
  2018.

\bibitem[\protect\citeauthoryear{Ho and Ermon}{2016}]{ho2016generative}
Jonathan Ho and Stefano Ermon.
\newblock Generative adversarial imitation learning.
\newblock In {\em Advances in Neural Information Processing Systems}, pages
  4565--4573, 2016.

\bibitem[\protect\citeauthoryear{Ho \bgroup \em et al.\egroup
  }{2016}]{ho2016showing}
Mark~K Ho, Michael Littman, James MacGlashan, Fiery Cushman, and Joseph~L
  Austerweil.
\newblock Showing versus doing: Teaching by demonstration.
\newblock In {\em Advances in neural information processing systems}, pages
  3027--3035, 2016.

\bibitem[\protect\citeauthoryear{Hussein \bgroup \em et al.\egroup
  }{2017}]{hussein2017imitation}
Ahmed Hussein, Mohamed~Medhat Gaber, Eyad Elyan, and Chrisina Jayne.
\newblock Imitation learning: A survey of learning methods.
\newblock {\em ACM Computing Surveys (CSUR)}, 50(2):21, 2017.

\bibitem[\protect\citeauthoryear{Ibarz \bgroup \em et al.\egroup
  }{2018}]{ibarz2018reward}
Borja Ibarz, Jan Leike, Tobias Pohlen, Geoffrey Irving, Shane Legg, and Dario
  Amodei.
\newblock Reward learning from human preferences and demonstrations in atari.
\newblock In {\em Advances in Neural Information Processing Systems}, pages
  8022--8034, 2018.

\bibitem[\protect\citeauthoryear{Isbell \bgroup \em et al.\egroup
  }{2001}]{isbell2001social}
Charles Isbell, Christian~R Shelton, Michael Kearns, Satinder Singh, and Peter
  Stone.
\newblock A social reinforcement learning agent.
\newblock In {\em Proceedings of the fifth international conference on
  Autonomous agents}, pages 377--384. ACM, 2001.

\bibitem[\protect\citeauthoryear{Kim \bgroup \em et al.\egroup
  }{2018}]{kim2018textual}
Jinkyu Kim, Anna Rohrbach, Trevor Darrell, John Canny, and Zeynep Akata.
\newblock Textual explanations for self-driving vehicles.
\newblock In {\em Proceedings of the European Conference on Computer Vision
  (ECCV)}, pages 563--578, 2018.

\bibitem[\protect\citeauthoryear{Knox and Stone}{2009}]{knox2009interactively}
W~Bradley Knox and Peter Stone.
\newblock Interactively shaping agents via human reinforcement: The tamer
  framework.
\newblock In {\em Proceedings of the fifth international conference on
  Knowledge capture}, pages 9--16. ACM, 2009.

\bibitem[\protect\citeauthoryear{Knox and Stone}{2010}]{knox2010combining}
W~Bradley Knox and Peter Stone.
\newblock Combining manual feedback with subsequent mdp reward signals for
  reinforcement learning.
\newblock In {\em Proceedings of the 9th International Conference on Autonomous
  Agents and Multiagent Systems: volume 1-Volume 1}, pages 5--12. International
  Foundation for Autonomous Agents and Multiagent Systems, 2010.

\bibitem[\protect\citeauthoryear{Knox and Stone}{2012}]{knox2012reinforcement}
W~Bradley Knox and Peter Stone.
\newblock Reinforcement learning from simultaneous human and mdp reward.
\newblock In {\em Proceedings of the 11th International Conference on
  Autonomous Agents and Multiagent Systems-Volume 1}, pages 475--482.
  International Foundation for Autonomous Agents and Multiagent Systems, 2012.

\bibitem[\protect\citeauthoryear{Le \bgroup \em et al.\egroup
  }{2018}]{le2018hierarchical}
Hoang Le, Nan Jiang, Alekh Agarwal, Miroslav Dudik, Yisong Yue, and Hal
  Daum{\'e}.
\newblock Hierarchical imitation and reinforcement learning.
\newblock In {\em International Conference on Machine Learning}, pages
  2923--2932, 2018.

\bibitem[\protect\citeauthoryear{Li \bgroup \em et al.\egroup
  }{2018}]{li2018eye}
Yin Li, Miao Liu, and James~M Rehg.
\newblock In the eye of beholder: Joint learning of gaze and actions in first
  person video.
\newblock In {\em Proceedings of the European Conference on Computer Vision
  (ECCV)}, pages 619--635, 2018.

\bibitem[\protect\citeauthoryear{Liu \bgroup \em et al.\egroup
  }{2018}]{liu2018imitation}
YuXuan Liu, Abhishek Gupta, Pieter Abbeel, and Sergey Levine.
\newblock Imitation from observation: Learning to imitate behaviors from raw
  video via context translation.
\newblock In {\em 2018 IEEE International Conference on Robotics and Automation
  (ICRA)}, pages 1118--1125. IEEE, 2018.

\bibitem[\protect\citeauthoryear{Loftin \bgroup \em et al.\egroup
  }{2016}]{loftin2016learning}
Robert Loftin, Bei Peng, James MacGlashan, Michael~L Littman, Matthew~E Taylor,
  Jeff Huang, and David~L Roberts.
\newblock Learning behaviors via human-delivered discrete feedback: modeling
  implicit feedback strategies to speed up learning.
\newblock {\em Autonomous agents and multi-agent systems}, 30(1):30--59, 2016.

\bibitem[\protect\citeauthoryear{MacGlashan \bgroup \em et al.\egroup
  }{2017}]{macglashan2017interactive}
James MacGlashan, Mark~K Ho, Robert Loftin, Bei Peng, Guan Wang, David~L
  Roberts, Matthew~E Taylor, and Michael~L Littman.
\newblock Interactive learning from policy-dependent human feedback.
\newblock In {\em Proceedings of the 34th International Conference on Machine
  Learning-Volume 70}, pages 2285--2294. JMLR. org, 2017.

\bibitem[\protect\citeauthoryear{Merel \bgroup \em et al.\egroup
  }{2017}]{merel2017learning}
Josh Merel, Yuval Tassa, Sriram Srinivasan, Jay Lemmon, Ziyu Wang, Greg Wayne,
  and Nicolas Heess.
\newblock Learning human behaviors from motion capture by adversarial
  imitation.
\newblock {\em arXiv preprint arXiv:1707.02201}, 2017.

\bibitem[\protect\citeauthoryear{Mnih \bgroup \em et al.\egroup
  }{2015}]{mnih2015human}
Volodymyr Mnih, Koray Kavukcuoglu, David Silver, Andrei~A Rusu, Joel Veness,
  Marc~G Bellemare, Alex Graves, Martin Riedmiller, Andreas~K Fidjeland, Georg
  Ostrovski, et~al.
\newblock Human-level control through deep reinforcement learning.
\newblock {\em Nature}, 518(7540):529--533, 2015.

\bibitem[\protect\citeauthoryear{Nair \bgroup \em et al.\egroup
  }{2017}]{nair2017combining}
Ashvin Nair, Dian Chen, Pulkit Agrawal, Phillip Isola, Pieter Abbeel, Jitendra
  Malik, and Sergey Levine.
\newblock Combining self-supervised learning and imitation for vision-based
  rope manipulation.
\newblock In {\em 2017 IEEE International Conference on Robotics and Automation
  (ICRA)}, pages 2146--2153. IEEE, 2017.

\bibitem[\protect\citeauthoryear{Osa \bgroup \em et al.\egroup
  }{2018}]{osa2018algorithmic}
Takayuki Osa, Joni Pajarinen, Gerhard Neumann, J~Andrew Bagnell, Pieter Abbeel,
  Jan Peters, et~al.
\newblock An algorithmic perspective on imitation learning.
\newblock {\em Foundations and Trends{\textregistered} in Robotics},
  7(1-2):1--179, 2018.

\bibitem[\protect\citeauthoryear{Palazzi \bgroup \em et al.\egroup
  }{2018}]{palazzi2018predicting}
Andrea Palazzi, Davide Abati, Simone Calderara, Francesco Solera, and Rita
  Cucchiara.
\newblock Predicting the driver's focus of attention: the dr (eye) ve project.
\newblock {\em IEEE transactions on pattern analysis and machine intelligence},
  2018.

\bibitem[\protect\citeauthoryear{Pilarski \bgroup \em et al.\egroup
  }{2011}]{pilarski2011online}
Patrick~M Pilarski, Michael~R Dawson, Thomas Degris, Farbod Fahimi, Jason~P
  Carey, and Richard~S Sutton.
\newblock Online human training of a myoelectric prosthesis controller via
  actor-critic reinforcement learning.
\newblock In {\em 2011 IEEE International Conference on Rehabilitation
  Robotics}, pages 1--7. IEEE, 2011.

\bibitem[\protect\citeauthoryear{Pinsler \bgroup \em et al.\egroup
  }{2018}]{pinsler2018sample}
Robert Pinsler, Riad Akrour, Takayuki Osa, Jan Peters, and Gerhard Neumann.
\newblock Sample and feedback efficient hierarchical reinforcement learning
  from human preferences.
\newblock In {\em 2018 IEEE International Conference on Robotics and Automation
  (ICRA)}, pages 596--601. IEEE, 2018.

\bibitem[\protect\citeauthoryear{Ross \bgroup \em et al.\egroup
  }{2011}]{ross2011reduction}
St{\'e}phane Ross, Geoffrey~J Gordon, and Drew Bagnell.
\newblock A reduction of imitation learning and structured prediction to
  no-regret online learning.
\newblock In {\em International Conference on Artificial Intelligence and
  Statistics}, pages 627--635, 2011.

\bibitem[\protect\citeauthoryear{Saunders \bgroup \em et al.\egroup
  }{2018}]{saunders2018trial}
William Saunders, Girish Sastry, Andreas Stuhlmueller, and Owain Evans.
\newblock Trial without error: Towards safe reinforcement learning via human
  intervention.
\newblock In {\em Proceedings of the 17th International Conference on
  Autonomous Agents and MultiAgent Systems}, pages 2067--2069. International
  Foundation for Autonomous Agents and Multiagent Systems, 2018.

\bibitem[\protect\citeauthoryear{Schaal}{1999}]{schaal1999imitation}
Stefan Schaal.
\newblock Is imitation learning the route to humanoid robots?
\newblock {\em Trends in cognitive sciences}, 3(6):233--242, 1999.

\bibitem[\protect\citeauthoryear{Sermanet \bgroup \em et al.\egroup
  }{2018}]{sermanet2018time}
Pierre Sermanet, Corey Lynch, Yevgen Chebotar, Jasmine Hsu, Eric Jang, Stefan
  Schaal, Sergey Levine, and Google Brain.
\newblock Time-contrastive networks: Self-supervised learning from video.
\newblock In {\em 2018 IEEE International Conference on Robotics and Automation
  (ICRA)}, pages 1134--1141. IEEE, 2018.

\bibitem[\protect\citeauthoryear{Silver \bgroup \em et al.\egroup
  }{2016}]{silver2016mastering}
David Silver, Aja Huang, Chris~J Maddison, Arthur Guez, Laurent Sifre, George
  Van Den~Driessche, Julian Schrittwieser, Ioannis Antonoglou, Veda
  Panneershelvam, Marc Lanctot, et~al.
\newblock Mastering the game of go with deep neural networks and tree search.
\newblock {\em Nature}, 529(7587):484--489, 2016.

\bibitem[\protect\citeauthoryear{Stadie \bgroup \em et al.\egroup
  }{2017}]{stadie2017third}
Bradly~C Stadie, Pieter Abbeel, and Ilya Sutskever.
\newblock Third-person imitation learning.
\newblock {\em International Conference on Learning Representations}, 2017.

\bibitem[\protect\citeauthoryear{Sutton and
  Barto}{2018}]{sutton2018reinforcement}
Richard~S Sutton and Andrew~G Barto.
\newblock {\em Reinforcement learning: An introduction}.
\newblock MIT press, 2018.

\bibitem[\protect\citeauthoryear{Sutton \bgroup \em et al.\egroup
  }{1999}]{sutton1999between}
Richard~S Sutton, Doina Precup, and Satinder Singh.
\newblock Between mdps and semi-mdps: A framework for temporal abstraction in
  reinforcement learning.
\newblock {\em Artificial intelligence}, 112(1-2):181--211, 1999.

\bibitem[\protect\citeauthoryear{Tenorio-Gonzalez \bgroup \em et al.\egroup
  }{2010}]{tenorio2010dynamic}
Ana~C Tenorio-Gonzalez, Eduardo~F Morales, and Luis Villase{\~n}or-Pineda.
\newblock Dynamic reward shaping: training a robot by voice.
\newblock In {\em Ibero-American conference on artificial intelligence}, pages
  483--492. Springer, 2010.

\bibitem[\protect\citeauthoryear{Thomaz and
  Breazeal}{2008}]{thomaz2008teachable}
Andrea~L Thomaz and Cynthia Breazeal.
\newblock Teachable robots: Understanding human teaching behavior to build more
  effective robot learners.
\newblock {\em Artificial Intelligence}, 172(6-7):716--737, 2008.

\bibitem[\protect\citeauthoryear{Todorov \bgroup \em et al.\egroup
  }{2012}]{todorov2012mujoco}
Emanuel Todorov, Tom Erez, and Yuval Tassa.
\newblock Mujoco: A physics engine for model-based control.
\newblock In {\em 2012 IEEE/RSJ International Conference on Intelligent Robots
  and Systems}, pages 5026--5033. IEEE, 2012.

\bibitem[\protect\citeauthoryear{Torabi \bgroup \em et al.\egroup
  }{2018a}]{torabi2018behavioral}
Faraz Torabi, Garrett Warnell, and Peter Stone.
\newblock Behavioral cloning from observation.
\newblock In {\em Proceedings of the 27th International Joint Conference on
  Artificial Intelligence}, pages 4950--4957. AAAI Press, 2018.

\bibitem[\protect\citeauthoryear{Torabi \bgroup \em et al.\egroup
  }{2018b}]{torabi2018generative}
Faraz Torabi, Garrett Warnell, and Peter Stone.
\newblock Generative adversarial imitation from observation.
\newblock {\em arXiv preprint arXiv:1807.06158}, 2018.

\bibitem[\protect\citeauthoryear{Torabi \bgroup \em et al.\egroup
  }{2019a}]{torabi2019adversarial}
Faraz Torabi, Garrett Warnell, and Peter Stone.
\newblock Adversarial imitation learning from state-only demonstrations.
\newblock In {\em Proceedings of the 18th International Conference on
  Autonomous Agents and MultiAgent Systems}, pages 2229--2231. International
  Foundation for Autonomous Agents and Multiagent Systems, 2019.

\bibitem[\protect\citeauthoryear{Torabi \bgroup \em et al.\egroup
  }{2019b}]{torabi2019imitation}
Faraz Torabi, Garrett Warnell, and Peter Stone.
\newblock Imitation learning from video by leveraging proprioception.
\newblock {\em arXiv preprint arXiv:1905.09335}, 2019.

\bibitem[\protect\citeauthoryear{Torabi \bgroup \em et al.\egroup
  }{2019c}]{torabi2019recent}
Faraz Torabi, Garrett Warnell, and Peter Stone.
\newblock Recent advances in imitation learning from observation.
\newblock {\em arXiv preprint arXiv:1905.13566}, 2019.

\bibitem[\protect\citeauthoryear{Warnell \bgroup \em et al.\egroup
  }{2018}]{warnell2018deep}
Garrett Warnell, Nicholas Waytowich, Vernon Lawhern, and Peter Stone.
\newblock Deep tamer: Interactive agent shaping in high-dimensional state
  spaces.
\newblock In {\em Thirty-Second AAAI Conference on Artificial Intelligence},
  2018.

\bibitem[\protect\citeauthoryear{Wilson \bgroup \em et al.\egroup
  }{2012}]{wilson2012bayesian}
Aaron Wilson, Alan Fern, and Prasad Tadepalli.
\newblock A bayesian approach for policy learning from trajectory preference
  queries.
\newblock In {\em Advances in neural information processing systems}, pages
  1133--1141, 2012.

\bibitem[\protect\citeauthoryear{Wirth \bgroup \em et al.\egroup
  }{2016}]{wirth2016model}
Christian Wirth, Johannes F{\"u}rnkranz, and Gerhard Neumann.
\newblock Model-free preference-based reinforcement learning.
\newblock In {\em Thirtieth AAAI Conference on Artificial Intelligence}, 2016.

\bibitem[\protect\citeauthoryear{Wirth \bgroup \em et al.\egroup
  }{2017}]{wirth2017survey}
Christian Wirth, Riad Akrour, Gerhard Neumann, and Johannes F{\"u}rnkranz.
\newblock A survey of preference-based reinforcement learning methods.
\newblock {\em The Journal of Machine Learning Research}, 18(1):4945--4990,
  2017.

\bibitem[\protect\citeauthoryear{Zhang \bgroup \em et al.\egroup
  }{2018}]{zhang2018agil}
Ruohan Zhang, Zhuode Liu, Luxin Zhang, Jake~A Whritner, Karl~S Muller, Mary~M
  Hayhoe, and Dana~H Ballard.
\newblock Agil: Learning attention from human for visuomotor tasks.
\newblock In {\em Proceedings of the European Conference on Computer Vision
  (ECCV)}, pages 663--679, 2018.

\bibitem[\protect\citeauthoryear{Zhang \bgroup \em et al.\egroup
  }{2019}]{zhang2019atari}
Ruohan Zhang, Zhuode Liu, Lin Guan, Luxin Zhang, Mary~M Hayhoe, and Dana~H
  Ballard.
\newblock Atari-head: Atari human eye-tracking and demonstration dataset.
\newblock {\em arXiv preprint arXiv:1903.06754}, 2019.

\bibitem[\protect\citeauthoryear{Zintgraf \bgroup \em et al.\egroup
  }{2018}]{zintgraf2018ordered}
Luisa~M Zintgraf, Diederik~M Roijers, Sjoerd Linders, Catholijn~M Jonker, and
  Ann Now{\'e}.
\newblock Ordered preference elicitation strategies for supporting
  multi-objective decision making.
\newblock In {\em Proceedings of the 17th International Conference on
  Autonomous Agents and MultiAgent Systems}, pages 1477--1485. International
  Foundation for Autonomous Agents and Multiagent Systems, 2018.

\end{thebibliography}
}
\end{document}